%% file: main.tex
\documentclass[10pt,twocolumn,letterpaper]{article}

\usepackage{cvpr}              

\usepackage{graphicx}
\usepackage{amsmath}
\usepackage{amssymb}
\usepackage{booktabs}
\usepackage{placeins}
\usepackage{float}
\usepackage{enumitem}
\usepackage{tabularx}
\usepackage{array}
\usepackage{ragged2e}
\usepackage{afterpage}
\usepackage{caption}
\usepackage{multirow}
\usepackage{xcolor}
\usepackage{algorithm}
\usepackage{algorithmic}
\usepackage{subcaption}

\usepackage[pagebackref,breaklinks,colorlinks]{hyperref}

\usepackage[capitalize]{cleveref}
\crefname{section}{Sec.}{Secs.}
\Crefname{section}{Section}{Sections}
\Crefname{table}{Table}{Tables}
\crefname{table}{Tab.}{Tabs.}

\def\cvprPaperID{*} 
\def\confName{CVPR}
\def\confYear{2025}

\begin{document}

\title{Dual-Channel Attention Guidance for Training-Free Image Editing Control in Diffusion Transformers}

\author{Guandong Li\\
iFLYTEK\\
\quad (Corresponding Author)\\
}

\twocolumn[{
\renewcommand\twocolumn[1][]{#1}
\maketitle
}]

\input{abstract}

\input{introduction}
\input{related_work}
\input{preliminary}
\input{method}
\input{experiments}
\input{discussion}

{\small
\bibliographystyle{ieee_fullname}
\bibliography{references}
}

\end{document}

%% file: abstract.tex
\begin{abstract}
Training-free control over editing intensity is a critical requirement for diffusion-based image editing models built on the Diffusion Transformer (DiT) architecture. Existing attention manipulation methods focus exclusively on the Key space to modulate attention routing, leaving the Value space—which governs feature aggregation—entirely unexploited. In this paper, we first reveal that both Key and Value projections in DiT's multi-modal attention layers exhibit a pronounced \emph{bias-delta} structure, where token embeddings cluster tightly around a layer-specific bias vector. Building on this observation, we propose \textbf{Dual-Channel Attention Guidance (DCAG)}, a training-free framework that simultaneously manipulates both the Key channel (controlling \emph{where} to attend) and the Value channel (controlling \emph{what} to aggregate). We provide a theoretical analysis showing that the Key channel operates through the nonlinear softmax function, acting as a coarse control knob, while the Value channel operates through linear weighted summation, serving as a fine-grained complement. Together, the two-dimensional parameter space $(\delta_k, \delta_v)$ enables more precise editing-fidelity trade-offs than any single-channel method. Extensive experiments on the PIE-Bench benchmark (700 images, 10 editing categories) demonstrate that DCAG consistently outperforms Key-only guidance across all fidelity metrics, with the most significant improvements observed in localized editing tasks such as object deletion ($\downarrow$4.9\% LPIPS) and object addition ($\downarrow$3.2\% LPIPS).
\end{abstract}

%% file: introduction.tex
\section{Introduction}
\label{sec:intro}

Recent advances in Diffusion Transformers (DiTs)~\cite{peebles2023scalable,ho2020denoising,rombach2022high} have enabled powerful instruction-guided image editing models such as Qwen-Image-Edit~\cite{qwen_image_edit} and Step1X-Edit~\cite{liu2025step1x}. These models accept a source image and a natural language editing instruction, producing an edited image that faithfully follows the instruction while preserving unrelated content. However, a fundamental challenge remains: \emph{how to precisely control the trade-off between editing strength and content preservation without additional training}.

Classifier-Free Guidance (CFG)~\cite{ho2022classifier} provides a basic mechanism for adjusting generation strength, but its control over editing intensity is coarse and often leads to artifacts at extreme scales. More recently, attention manipulation methods have emerged as a promising direction~\cite{hertz2022prompt,chefer2023attend,li2024layout}. GRAG~\cite{zhang2025group} decomposes the Key projections in multi-modal attention layers into a bias and delta component, then rescales them to control editing intensity. While effective, GRAG and similar approaches operate exclusively on the Key space, which governs \emph{attention routing}—determining which tokens receive attention. The Value space, which controls \emph{what content is aggregated} after attention weights are computed, remains entirely unexploited.

In this work, we make the following key observation: the bias-delta structure is not unique to Keys. \textbf{Value projections in DiT's multi-modal attention layers exhibit the same pronounced clustering phenomenon}, where all token embeddings concentrate around a layer-specific bias vector. This suggests that the Value space offers an independent, orthogonal channel for editing control.

Building on this observation, we propose \textbf{Dual-Channel Attention Guidance (DCAG)}, a training-free framework that simultaneously manipulates both the Key and Value channels. The attention output in a transformer layer can be decomposed as:
\begin{equation}
    \text{Output} = \underbrace{\text{softmax}\!\left(\frac{QK^\top}{\sqrt{d}}\right)}_{\text{K: attention routing}} \cdot \underbrace{V}_{\text{V: feature aggregation}}
\end{equation}
DCAG applies bias-delta rescaling to both $K$ and $V$, creating a two-dimensional parameter space $(\delta_k, \delta_v)$ that enables finer-grained control than any single-channel method. We provide a theoretical analysis revealing that the two channels have fundamentally different control characteristics: the Key channel operates through the nonlinear softmax function, where small perturbations are exponentially amplified (coarse control), while the Value channel operates through linear weighted summation, producing proportional and predictable effects (fine control).

Our contributions are summarized as follows:
\begin{itemize}
    \item We discover that both Key and Value spaces in DiT's multi-modal attention exhibit a bias-delta structure, revealing the Value space as an overlooked control channel for editing guidance.
    \item We provide a theoretical analysis of the distinct control characteristics of the Key channel (nonlinear, dominant) and Value channel (linear, complementary), explaining their orthogonal roles.
    \item We propose DCAG, a unified dual-channel framework with a 2D parameter space $(\delta_k, \delta_v)$ that subsumes single-channel methods as special cases.
    \item We conduct extensive experiments on PIE-Bench, demonstrating consistent improvements over Key-only guidance across 8 out of 10 editing categories, with up to 4.9\% LPIPS reduction in localized editing tasks.
\end{itemize}

%% file: related_work.tex
\section{Related Work}
\label{sec:related}

\paragraph{Training-Free Editing Control.}
Controlling the intensity of image edits without retraining is a long-standing challenge. Classifier-Free Guidance (CFG)~\cite{ho2022classifier} adjusts the balance between conditional and unconditional predictions during sampling, but provides only coarse control over editing strength~\cite{zhang2025group}. SDEdit~\cite{meng2021sdedit} controls editing intensity through the noise level added to the source image, but conflates structural changes with detail preservation. Null-text Inversion~\cite{mokady2023null} enables faithful reconstruction via DDIM inversion~\cite{song2020denoising} for subsequent editing, while Plug-and-Play~\cite{tumanyan2023plug} leverages diffusion features for image-to-image translation. InstructPix2Pix~\cite{brooks2023instructpix2pix} trains an end-to-end model for instruction-based editing, and Imagic~\cite{kawar2023imagic} optimizes text embeddings for real image editing. FlowEdit~\cite{kulikov2025flowedit} and related flow-based methods~\cite{lipman2022flow,liu2022flow} offer alternative control mechanisms but require careful tuning of multiple hyperparameters. In contrast, our approach operates directly within the attention mechanism, providing interpretable and orthogonal control through two distinct channels.

\paragraph{Attention Manipulation for Image Editing.}
Manipulating attention maps has proven effective for controlling diffusion-based generation and editing. Prompt-to-Prompt (P2P)~\cite{hertz2022prompt} injects cross-attention maps from the source generation into the editing process to preserve spatial layout. Attend-and-Excite~\cite{chefer2023attend} optimizes latents to ensure all subjects receive sufficient attention. Self-Guidance~\cite{epstein2023diffusion} leverages internal attention features for spatial control. MasaCtrl~\cite{cao2023masactrl} performs mutual self-attention control for consistent image synthesis. Perturbed-Attention Guidance (PAG)~\cite{ahn2024self} improves sample quality by perturbing self-attention. Prior work on attention-based layout control~\cite{li2024layout} has also demonstrated the effectiveness of manipulating attention losses for spatial guidance. Most recently, GRAG~\cite{zhang2025group} decomposes Key projections into bias and delta components, rescaling them to control editing intensity. However, all these methods focus on either the Query-Key interaction (attention weights) or cross-attention maps, without exploiting the Value space. Our work is the first to identify and leverage the bias-delta structure in Value projections as a complementary control channel.

\paragraph{Diffusion Transformers.}
The Diffusion Transformer (DiT) architecture~\cite{peebles2023scalable} replaces the U-Net backbone~\cite{ho2020denoising,dhariwal2021diffusion,rombach2022high} with transformer blocks, enabling better scalability and performance. Subsequent work has scaled DiTs with rectified flow objectives~\cite{esser2024scaling,liu2022flow} and improved latent representations~\cite{podell2023sdxl}. Modern DiT-based editing models such as Qwen-Image-Edit~\cite{qwen_image_edit} employ dual-stream architectures where text and image tokens are processed through joint multi-modal attention (MM-Attention) layers. In these architectures, the Key and Value projections of image tokens carry rich structural information about the source image, making them natural targets for editing control. Our DCAG framework is designed for this dual-stream MM-Attention setting but is applicable to any transformer-based diffusion model.

%% file: preliminary.tex
\section{Preliminary}
\label{sec:preliminary}

\subsection{Multi-Modal Attention in DiT}

Modern DiT-based image editing models~\cite{peebles2023scalable,esser2024scaling} employ a dual-stream architecture with $L$ transformer blocks (typically $L=60$). Each block contains a multi-modal attention (MM-Attention) layer that jointly processes text tokens $\mathbf{X}_\text{txt} \in \mathbb{R}^{S_t \times D}$ and image tokens $\mathbf{X}_\text{img} \in \mathbb{R}^{S_i \times D}$, where $S_t$ and $S_i$ are the sequence lengths and $D$ is the hidden dimension.

For each stream, separate Query, Key, and Value projections are computed:
\begin{align}
    Q_s &= W_Q^s \mathbf{X}_s, \quad K_s = W_K^s \mathbf{X}_s, \nonumber \\
    V_s &= W_V^s \mathbf{X}_s, \quad s \in \{\text{txt}, \text{img}\}
\end{align}
After applying rotary position embeddings (RoPE)~\cite{su2024roformer} to $Q$ and $K$, the streams are concatenated for joint attention:
\begin{equation}
    Q = [Q_\text{txt}; Q_\text{img}], \quad K = [K_\text{txt}; K_\text{img}], \quad V = [V_\text{txt}; V_\text{img}]
\end{equation}
\begin{equation}
    \text{Output} = \text{softmax}\!\left(\frac{QK^\top}{\sqrt{d_h}}\right) V
    \label{eq:attention}
\end{equation}
where $d_h$ is the per-head dimension. The output is then split back into text and image streams for subsequent processing.

\subsection{The Bias-Delta Structure}

A key empirical observation, first noted for Key projections~\cite{zhang2025group}, is that token embeddings in MM-Attention layers exhibit a pronounced clustering pattern. For the image Key tokens $\{K_\text{img}^i\}_{i=1}^{S_i}$ at any given layer, all tokens concentrate tightly around a shared bias vector:
\begin{equation}
    K_\text{img}^i = \underbrace{\bar{K}_\text{img}}_{\text{bias}} + \underbrace{(K_\text{img}^i - \bar{K}_\text{img})}_{\text{delta } \Delta K^i}
    \label{eq:bias_delta_k}
\end{equation}
where $\bar{K}_\text{img} = \frac{1}{S_i}\sum_{i=1}^{S_i} K_\text{img}^i$ is the token-wise mean. The bias component $\bar{K}_\text{img}$ encodes the layer's overall editing behavior, while the delta component $\Delta K^i$ captures token-specific content signals.

\textbf{Our key finding:} We discover that this bias-delta structure is \emph{not unique to Keys}. The Value projections exhibit the same clustering phenomenon:
\begin{equation}
    V_\text{img}^i = \underbrace{\bar{V}_\text{img}}_{\text{bias}} + \underbrace{(V_\text{img}^i - \bar{V}_\text{img})}_{\text{delta } \Delta V^i}
    \label{eq:bias_delta_v}
\end{equation}
We provide empirical evidence for this observation through systematic profiling across all 60 layers and 24 denoising steps in \cref{sec:v_profiling}. This discovery motivates our dual-channel approach: if both $K$ and $V$ have exploitable bias-delta structure, manipulating both channels should provide richer control than manipulating either alone.

%% file: method.tex
\section{Method: Dual-Channel Attention Guidance}
\label{sec:method}

\begin{figure}[t]
\centering
\includegraphics[width=\linewidth]{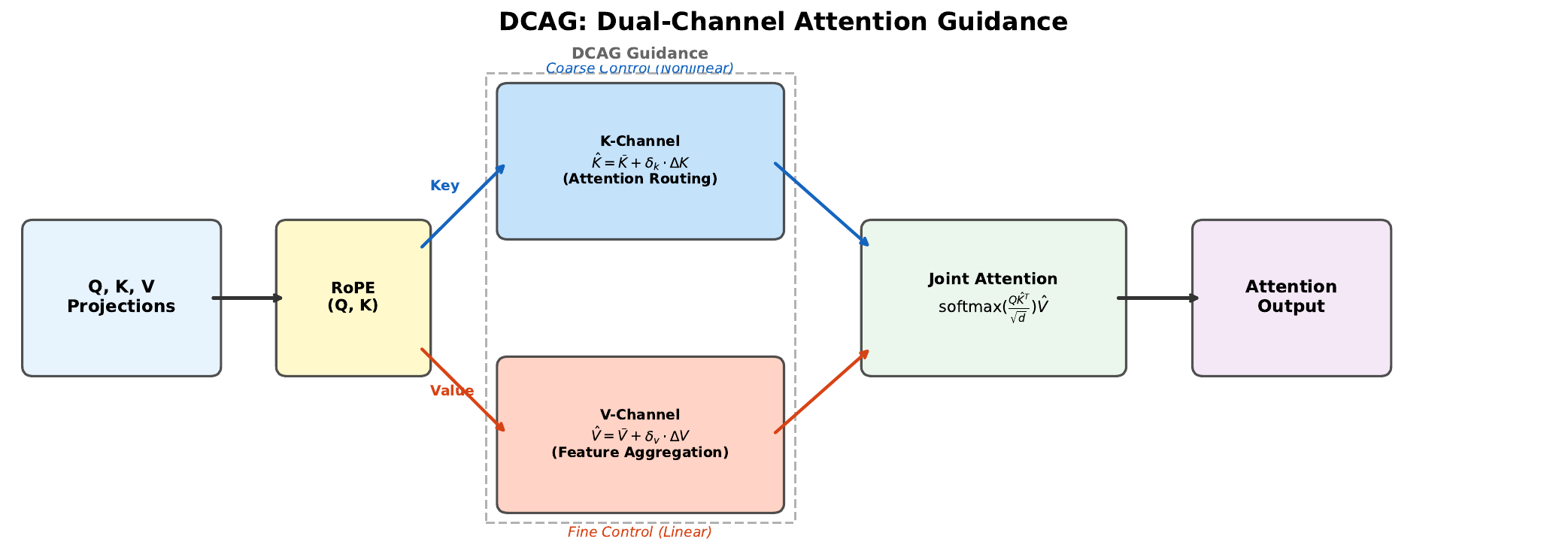}
\caption{Overview of DCAG. The Key channel controls attention routing (coarse, nonlinear) and the Value channel controls feature aggregation (fine, linear). Both channels apply bias-delta rescaling independently before the joint attention computation.}
\label{fig:method}
\end{figure}

\cref{fig:method} illustrates the overall DCAG framework. After RoPE encoding, DCAG independently rescales the Key and Value projections before the joint attention computation.

\subsection{Dual-Channel Rescaling}
\label{sec:rescaling}

Given the bias-delta decomposition of both Key and Value projections (\cref{eq:bias_delta_k,eq:bias_delta_v}), DCAG applies independent rescaling to each channel. For the image tokens at layer $l$:

\paragraph{Key Channel (Attention Routing).}
\begin{equation}
    \hat{K}_\text{img}^i = \lambda_k \cdot \bar{K}_\text{img} + \delta_k \cdot \Delta K^i
    \label{eq:k_rescale}
\end{equation}

\paragraph{Value Channel (Feature Aggregation).}
\begin{equation}
    \hat{V}_\text{img}^i = \lambda_v \cdot \bar{V}_\text{img} + \delta_v \cdot \Delta V^i
    \label{eq:v_rescale}
\end{equation}

where $(\lambda_k, \delta_k)$ and $(\lambda_v, \delta_v)$ are the bias and delta scales for each channel. Following prior findings that the delta scale is the primary control parameter~\cite{zhang2025group}, we fix $\lambda_k = \lambda_v = 1.0$ and focus on the delta scales $(\delta_k, \delta_v)$ as the two control knobs. When $\delta_k = \delta_v = 1.0$, DCAG reduces to the identity (no guidance). When $\delta_v = 1.0$, DCAG reduces to Key-only guidance.

The rescaling is applied after RoPE and before the joint attention computation (\cref{eq:attention}), preserving positional encoding integrity. The complete procedure is summarized in \cref{alg:dcag}.

\begin{algorithm}[t]
\caption{Dual-Channel Attention Guidance (DCAG)}
\label{alg:dcag}
\begin{algorithmic}[1]
\REQUIRE $Q, K, V \in \mathbb{R}^{B \times S \times H \times D}$; image token range $[i_s, i_e]$; scales $\delta_k, \delta_v$
\STATE $Q, K \leftarrow \text{RoPE}(Q), \text{RoPE}(K)$
\STATE \textcolor{blue}{\texttt{// Key channel}}
\STATE $\bar{K} \leftarrow \text{mean}(K[:, i_s\!:\!i_e], \text{dim}=1)$
\STATE $K[:, i_s\!:\!i_e] \leftarrow \bar{K} + \delta_k \cdot (K[:, i_s\!:\!i_e] - \bar{K})$
\STATE \textcolor{blue}{\texttt{// Value channel}}
\STATE $\bar{V} \leftarrow \text{mean}(V[:, i_s\!:\!i_e], \text{dim}=1)$
\STATE $V[:, i_s\!:\!i_e] \leftarrow \bar{V} + \delta_v \cdot (V[:, i_s\!:\!i_e] - \bar{V})$
\STATE $\hat{A} \leftarrow \text{Attention}(Q, K, V)$
\RETURN $\hat{A}$
\end{algorithmic}
\end{algorithm}

\subsection{Theoretical Analysis: Key vs.\ Value Channels}
\label{sec:theory}

To understand why the two channels provide complementary control, we analyze how perturbations in $\delta_k$ and $\delta_v$ affect the attention output. Consider a single query token $q$ attending to image tokens. The attention output is:
\begin{equation}
    o = \sum_i \alpha_i \hat{V}^i, \; \alpha_i = \frac{\exp(q^\top \hat{K}^i / \sqrt{d_h})}{\sum_j \exp(q^\top \hat{K}^j / \sqrt{d_h})}
\end{equation}

\paragraph{Key Channel: Nonlinear Amplification.}
Substituting the Key rescaling (\cref{eq:k_rescale}) with $\lambda_k = 1$:
\begin{equation}
    q^\top \hat{K}^i = q^\top \bar{K} + \delta_k \cdot q^\top \Delta K^i
\end{equation}
The first term $q^\top \bar{K}$ is shared across all tokens and cancels in the softmax normalization. The effective attention logit difference between tokens $i$ and $j$ is:
\begin{equation}
    q^\top \hat{K}^i - q^\top \hat{K}^j = \delta_k \cdot q^\top (\Delta K^i - \Delta K^j)
\end{equation}
Through the softmax's exponential function, this linear scaling of logit differences produces a \emph{nonlinear, amplified} effect on the attention distribution. Increasing $\delta_k$ sharpens the attention distribution, making it more peaked on tokens with high relevance. This is a \textbf{coarse control} mechanism—small changes in $\delta_k$ can dramatically redistribute attention weights.

\paragraph{Value Channel: Linear Proportionality.}
Substituting the Value rescaling (\cref{eq:v_rescale}) with $\lambda_v = 1$:
\begin{align}
    o &= \sum_i \alpha_i (\bar{V} + \delta_v \cdot \Delta V^i) \nonumber \\
      &= \bar{V} + \delta_v \sum_i \alpha_i \Delta V^i
\end{align}
The output decomposes into a fixed bias term $\bar{V}$ (independent of $\delta_v$) and a scaled deviation term. The effect of $\delta_v$ on the output is \emph{strictly linear}: doubling $\delta_v$ doubles the deviation from the mean. This is a \textbf{fine control} mechanism—changes in $\delta_v$ produce proportional, predictable effects on the output features.

\paragraph{Orthogonality.}
The Key channel modifies the attention weights $\{\alpha_i\}$ (which tokens are attended to), while the Value channel modifies the features $\{V^i\}$ (what content is aggregated). These operate on different factors of the attention output $o = \sum_i \alpha_i V^i$, making them functionally orthogonal. Increasing $\delta_v$ amplifies per-token feature distinctiveness without changing the attention distribution, thereby preserving local details in non-edited regions.

\subsection{2D Parameter Space}
\label{sec:2d_space}

DCAG's two control knobs $(\delta_k, \delta_v)$ span a 2D parameter plane. Each point in this plane corresponds to a specific editing-fidelity trade-off:

\begin{itemize}
    \item \textbf{Origin} $(1.0, 1.0)$: No guidance (baseline model behavior).
    \item \textbf{K-axis} $(\delta_k > 1, 1.0)$: Key-only guidance (equivalent to prior methods).
    \item \textbf{V-axis} $(1.0, \delta_v > 1)$: Value-only guidance (novel).
    \item \textbf{Interior} $(\delta_k > 1, \delta_v > 1)$: Dual-channel guidance (DCAG).
\end{itemize}

A key property of this 2D space is the existence of \emph{iso-fidelity contours}—curves along which the fidelity metric (e.g., LPIPS) remains approximately constant. Moving along such a contour trades off between Key-channel and Value-channel contributions while maintaining the same overall preservation level. This enables practitioners to find operating points that optimize editing quality at a desired fidelity level, which is impossible with single-channel methods.

%% file: experiments.tex
\section{Experiments}
\label{sec:experiments}

\subsection{Experimental Setup}

\paragraph{Benchmark.} We evaluate on PIE-Bench~\cite{ju2023pnp}, a comprehensive image editing benchmark containing 700 images across 10 editing categories: Random, Change Object, Add Object, Delete Object, Change Attribute, Change Count, Change Background, Change Style, Change Action, and Change Position.

\paragraph{Metrics.} We report four fidelity metrics measuring content preservation between source and edited images: LPIPS$\downarrow$~\cite{zhang2018unreasonable} (perceptual distance), SSIM$\uparrow$~\cite{wang2004image} (structural similarity), PSNR$\uparrow$ (peak signal-to-noise ratio), and MSE$\downarrow$. We additionally report CLIP-Score$\uparrow$~\cite{radford2021learning} to measure editing quality (semantic alignment between the edited image and the editing instruction).

\paragraph{Model.} All experiments use Qwen-Image-Edit~\cite{qwen_image_edit}, a 60-layer dual-stream DiT model. Inference uses 24 denoising steps, CFG scale 4.0, image resolution $1024 \times 1024$, and seed 42 for reproducibility.

\paragraph{Baselines.} We compare against: (1) \textbf{No Guidance}: the base model without any attention manipulation ($\delta_k = \delta_v = 1.0$); (2) \textbf{GRAG}~\cite{zhang2025group}: the state-of-the-art Key-only attention guidance method, which rescales Key deltas with $\delta_k > 1$ while keeping $\delta_v = 1.0$. We evaluate GRAG at multiple $\delta_k$ values to ensure fair comparison.

\subsection{Main Results}

\cref{tab:main} presents the main comparison between DCAG and GRAG on the full PIE-Bench (700 images).

\begin{table}[t]
\centering
\small
\caption{Main results on PIE-Bench (700 images). DCAG consistently outperforms GRAG (K-only) at matched $\delta_k$. Best results per group in \textbf{bold}.}
\label{tab:main}
\setlength{\tabcolsep}{3pt}
\begin{tabular}{lcccccc}
\toprule
Method & $\delta_k$ & $\delta_v$ & LPIPS$\downarrow$ & SSIM$\uparrow$ & PSNR$\uparrow$ & MSE$\downarrow$ \\
\midrule
No Guidance & 1.00 & 1.00 & 0.3523 & 0.6307 & 15.56 & 3902 \\
\midrule
\multicolumn{7}{l}{\textit{$\delta_k = 1.10$ group}} \\
GRAG & 1.10 & 1.00 & 0.2588 & 0.7444 & \textbf{17.93} & 2588 \\
DCAG & 1.10 & 1.05 & 0.2575 & 0.7453 & 17.92 & 2585 \\
DCAG & 1.10 & 1.10 & 0.2555 & 0.7468 & 17.90 & 2570 \\
DCAG & 1.10 & 1.15 & \textbf{0.2542} & \textbf{0.7477} & 17.89 & \textbf{2557} \\
DCAG & 1.10 & 1.20 & 0.2546 & 0.7466 & 17.81 & 2563 \\
\midrule
\multicolumn{7}{l}{\textit{$\delta_k = 1.15$ group}} \\
GRAG & 1.15 & 1.00 & 0.1991 & \textbf{0.8066} & \textbf{19.81} & 1751 \\
DCAG & 1.15 & 1.15 & \textbf{0.1974} & 0.8053 & 19.60 & \textbf{1742} \\
\midrule
\multicolumn{7}{l}{\textit{$\delta_k = 1.20$ group}} \\
GRAG & 1.20 & 1.00 & 0.1398 & 0.8557 & 21.91 & 1060 \\
\midrule
\multicolumn{7}{l}{\textit{Control: V-only cannot replace K}} \\
DCAG & 1.05 & 1.10 & 0.3126 & 0.6744 & 16.24 & 3497 \\
\bottomrule
\end{tabular}
\end{table}


Key observations:
\begin{itemize}
    \item \textbf{Attention guidance dramatically improves fidelity.} The No Guidance baseline (LPIPS = 0.3523) confirms that the base model produces significant unintended modifications. GRAG ($\delta_k\!=\!1.10$) reduces LPIPS by 26.5\% to 0.2588, and DCAG ($\delta_k\!=\!1.10, \delta_v\!=\!1.15$) further reduces it by 27.8\% to 0.2542.
    \item \textbf{DCAG improves upon K-only guidance.} At $\delta_k = 1.10$, adding Value-channel guidance monotonically improves LPIPS from 0.2588 ($\delta_v\!=\!1.0$) to 0.2542 ($\delta_v\!=\!1.15$), a 1.8\% reduction. The sweet spot is $\delta_v = 1.15$; further increasing to $\delta_v = 1.20$ causes slight regression (0.2546), indicating a saturation effect.
    \item \textbf{V-channel effect is near-monotonic with saturation.} Across the range $\delta_v \in [1.0, 1.15]$, all four fidelity metrics improve monotonically. Beyond 1.15, LPIPS and SSIM slightly regress while PSNR continues to decrease, suggesting the Value-channel amplification begins to distort fine-grained features.
    \item \textbf{K-channel is dominant.} Reducing $\delta_k$ from 1.10 to 1.05 (with $\delta_v = 1.10$) causes a 20.8\% LPIPS degradation, far exceeding the 1.8\% improvement from the full V-channel range. This confirms the theoretical prediction: the Key channel (nonlinear softmax amplification) has a much stronger effect than the Value channel (linear scaling).
    \item \textbf{Diminishing V-channel returns at higher $\delta_k$.} At $\delta_k = 1.15$, adding $\delta_v = 1.15$ yields only 0.9\% LPIPS improvement (0.1991$\to$0.1974), with SSIM and PSNR slightly decreasing. This suggests that stronger Key-channel guidance already captures most of the achievable fidelity gain, leaving less room for Value-channel contribution.
\end{itemize}

\subsection{Per-Category Analysis}

\cref{tab:per_type} shows the per-category breakdown comparing the best DCAG configuration ($\delta_k\!=\!1.10, \delta_v\!=\!1.15$) against GRAG ($\delta_k\!=\!1.10$).

\begin{table}[t]
\centering
\caption{Per-category LPIPS$\downarrow$ comparison at $\delta_k\!=\!1.10$. DCAG ($\delta_v\!=\!1.15$) improves 8 out of 10 categories.}
\label{tab:per_type}
\small
\begin{tabular}{lcccr}
\toprule
Category & GRAG & DCAG & $\Delta$ & $N$ \\
\midrule
Random & 0.2676 & 0.2594 & $\downarrow$3.1\% & 140 \\
Change Object & 0.2364 & 0.2299 & $\downarrow$2.7\% & 80 \\
Add Object & 0.1097 & 0.1068 & $\downarrow$2.7\% & 80 \\
Delete Object & 0.1752 & 0.1677 & $\downarrow$\textbf{4.3\%} & 80 \\
Change Attribute & 0.1538 & 0.1554 & $\uparrow$1.0\% & 40 \\
Change Count & 0.2688 & 0.2654 & $\downarrow$1.3\% & 40 \\
Change Background & 0.1564 & 0.1498 & $\downarrow$4.2\% & 40 \\
Change Style & 0.2020 & 0.1944 & $\downarrow$3.7\% & 40 \\
Change Action & 0.4290 & 0.4235 & $\downarrow$1.3\% & 80 \\
Change Position & 0.4554 & 0.4597 & $\uparrow$0.9\% & 80 \\
\midrule
\textbf{Overall} & \textbf{0.2588} & \textbf{0.2542} & $\downarrow$\textbf{1.8\%} & \textbf{700} \\
\bottomrule
\end{tabular}
\end{table}

The improvement pattern reveals an important insight about the Value channel's operating characteristics:

\textbf{Broad improvement at moderate $\delta_k$.} At $\delta_k = 1.10$, DCAG improves 8 out of 10 categories. The largest gains appear in Delete Object ($\downarrow$4.3\%), Change Background ($\downarrow$4.2\%), and Change Style ($\downarrow$3.7\%). These categories involve either localized editing regions or texture-level changes where amplifying per-token Value distinctiveness directly reduces feature mixing in non-edited areas.

\textbf{Interaction with $\delta_k$ level.} At $\delta_k = 1.15$, the per-category pattern shifts: Random ($\downarrow$3.0\%), Change Object ($\downarrow$3.1\%), and Change Position ($\downarrow$2.4\%) still improve, but Add Object ($\uparrow$11.2\%) and Change Style ($\uparrow$7.6\%) regress. This suggests that at higher Key-channel guidance, the attention routing already strongly preserves non-edited regions, and additional Value amplification can over-sharpen features in categories with small editing footprints.

\textbf{Practical implication.} The Value channel is most effective as a complement to moderate Key-channel guidance ($\delta_k \approx 1.10$), where it provides consistent improvement across nearly all categories. At stronger Key guidance, the Value channel should be used conservatively or omitted.

\subsection{Ablation: V-Channel Saturation}
\label{sec:ablation}

\cref{tab:v_sweep} shows the effect of sweeping $\delta_v$ at fixed $\delta_k = 1.10$, revealing the Value channel's saturation behavior.

\begin{table}[t]
\centering
\caption{V-channel sweep at $\delta_k = 1.10$. LPIPS improves monotonically up to $\delta_v = 1.15$, then saturates.}
\label{tab:v_sweep}
\begin{tabular}{ccccc}
\toprule
$\delta_v$ & LPIPS$\downarrow$ & SSIM$\uparrow$ & PSNR$\uparrow$ & MSE$\downarrow$ \\
\midrule
1.00 & 0.2588 & 0.7444 & 17.93 & 2588 \\
1.05 & 0.2575 & 0.7453 & 17.92 & 2585 \\
1.10 & 0.2555 & 0.7468 & 17.90 & 2570 \\
1.15 & \textbf{0.2542} & \textbf{0.7477} & 17.89 & \textbf{2557} \\
1.20 & 0.2546 & 0.7466 & 17.81 & 2563 \\
\bottomrule
\end{tabular}
\end{table}

The saturation at $\delta_v \approx 1.15$ is consistent with the linear nature of the Value channel: beyond a certain amplification, the per-token deviations $\Delta V^i$ become large enough to distort the feature content rather than sharpen it. This contrasts with the Key channel, where the nonlinear softmax amplification allows a wider effective range before saturation.

\begin{figure}[t]
\centering
\includegraphics[width=0.75\linewidth]{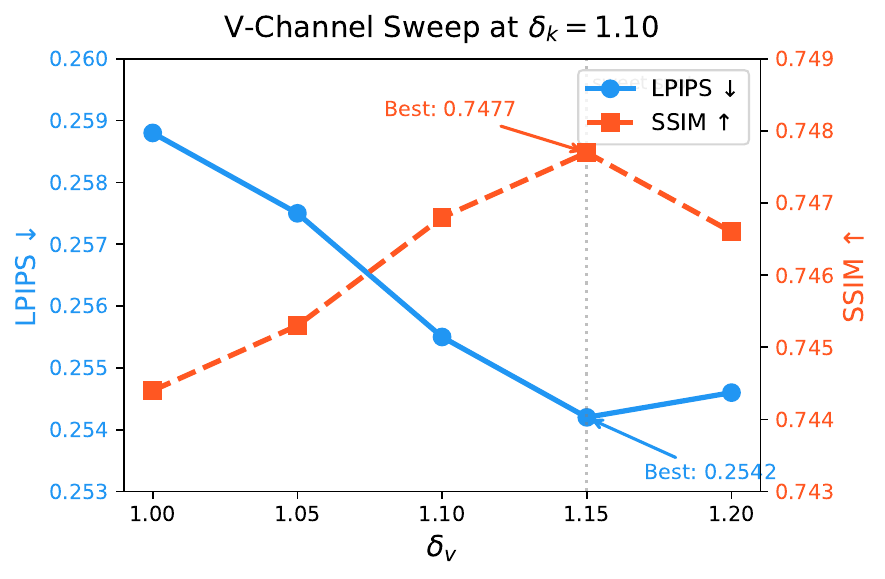}
\caption{V-channel sweep at fixed $\delta_k = 1.10$. LPIPS (blue, left axis) improves monotonically up to $\delta_v = 1.15$, then saturates. SSIM (red, right axis) follows the same pattern. The sweet spot at $\delta_v \approx 1.15$ reflects the linear nature of the Value channel.}
\label{fig:v_sweep}
\end{figure}

\begin{figure}[t]
\centering
\includegraphics[width=\linewidth]{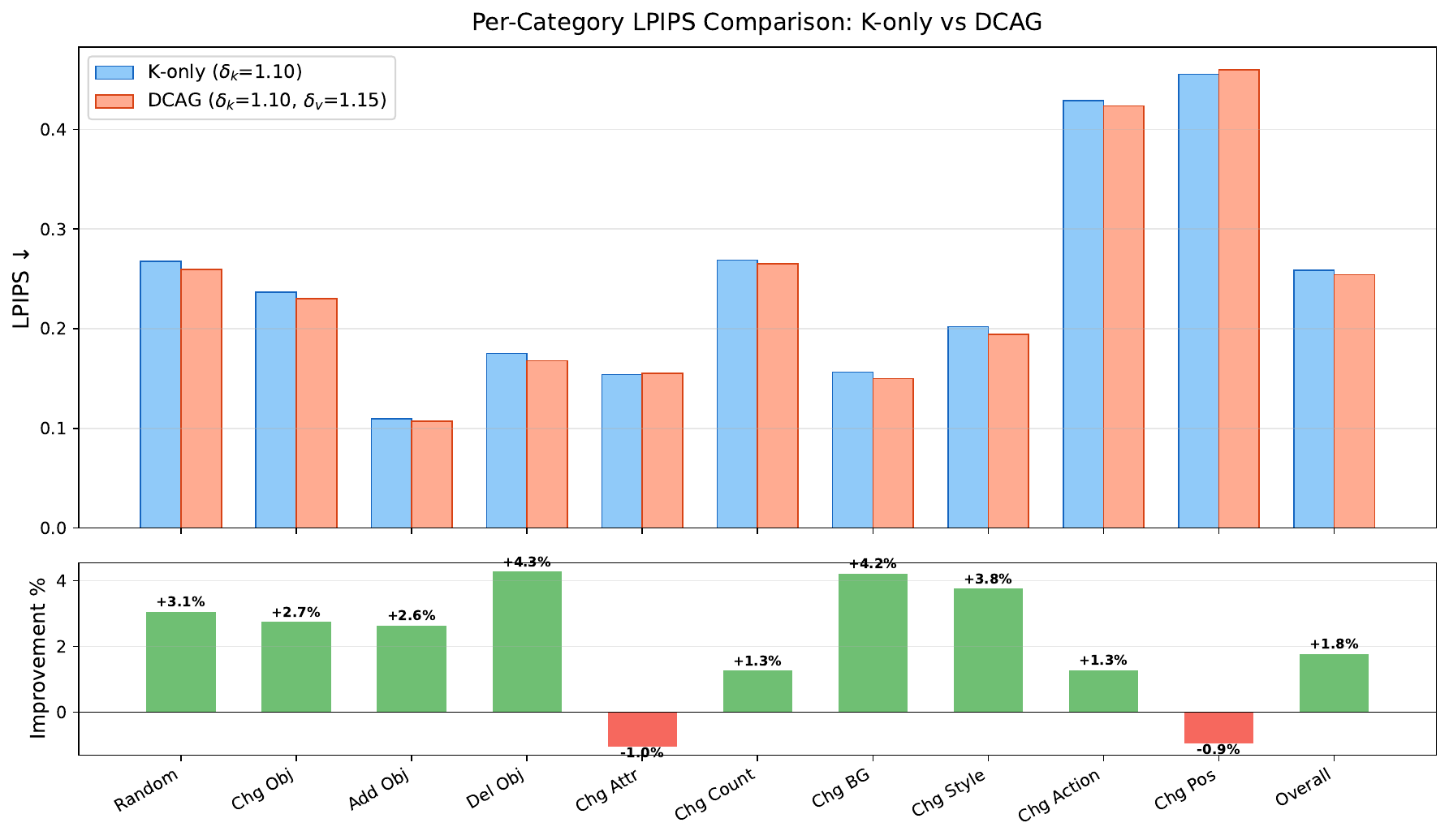}
\caption{Per-category LPIPS comparison between K-only ($\delta_k\!=\!1.10$) and DCAG ($\delta_k\!=\!1.10, \delta_v\!=\!1.15$). Top: absolute LPIPS values. Bottom: relative improvement (\%). DCAG improves 8 out of 10 categories, with the largest gains in Delete Object ($\downarrow$4.3\%) and Change Background ($\downarrow$4.2\%).}
\label{fig:per_category}
\end{figure}

\subsection{Value-Space Profiling}
\label{sec:v_profiling}

To empirically validate that the bias-delta structure exists in the Value space, we profile both Key and Value projections across all 60 layers and 24 denoising steps. For each space $X \in \{K, V\}$, layer $l$, and step $t$, we compute the delta-to-bias ratio:
\begin{equation}
    r_X^{(l,t)} = \frac{\text{mean}_i \|X^i - \bar{X}\|_2}{\|\bar{X}\|_2}
\end{equation}

\begin{figure}[t]
\centering
\includegraphics[width=\linewidth]{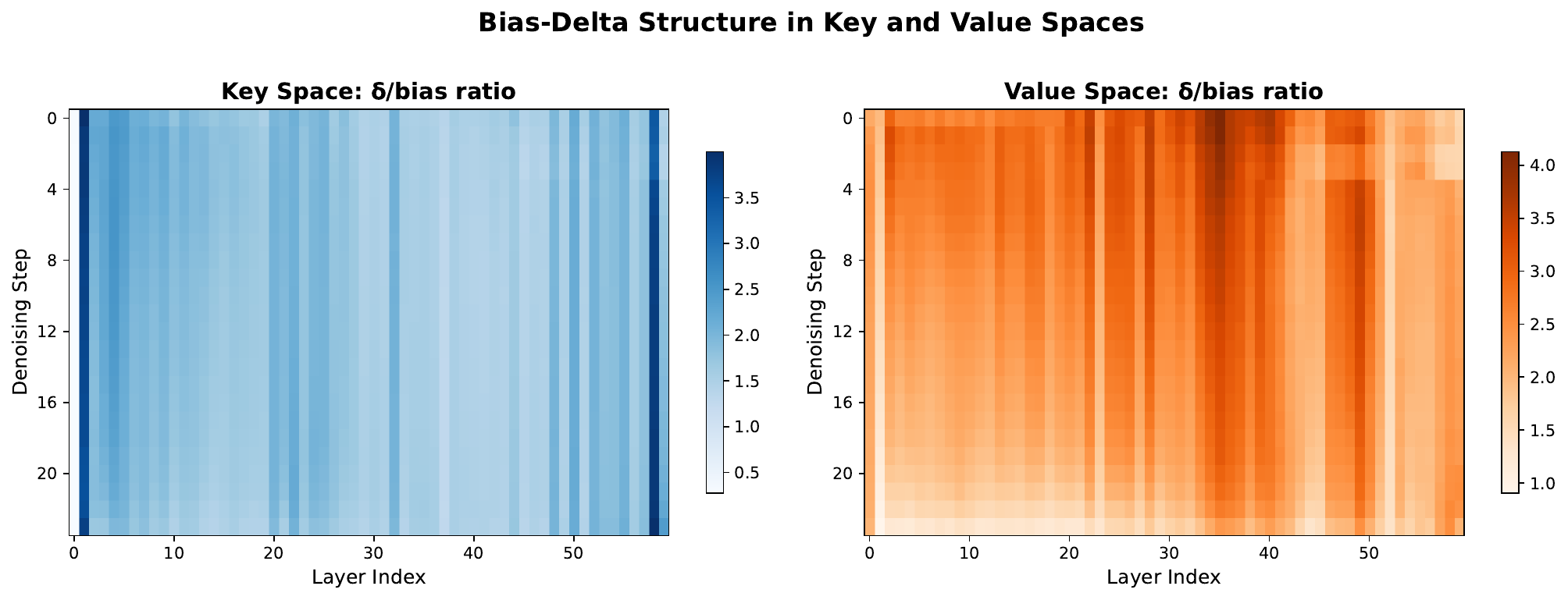}
\caption{Delta-to-bias ratio heatmaps across 60 layers (x-axis) and 24 denoising steps (y-axis). Left: Key space. Right: Value space. Both spaces exhibit pervasive bias-delta structure, with Value ratios (mean 2.45) consistently higher than Key ratios (mean 1.79). The weak correlation ($r = -0.17$) between the two confirms their structural independence.}
\label{fig:kv_profiling}
\end{figure}

\cref{fig:kv_profiling} presents the ratio heatmaps for both spaces. The results provide strong evidence for our hypothesis:

\begin{itemize}
    \item \textbf{Value space exhibits pervasive delta structure.} The delta-to-bias ratio is significantly greater than zero across all 1440 layer-step combinations (100\%), with a mean ratio of 2.45. This confirms that Value tokens cluster around a layer-specific bias with meaningful per-token deviations—the prerequisite for our rescaling approach.
    \item \textbf{Value deltas are relatively stronger than Key deltas.} The mean V-ratio (2.45) exceeds the mean K-ratio (1.79) by a factor of 1.37$\times$, indicating that Value space actually has proportionally larger deviations from the bias. This suggests Value-space manipulation has substantial room for effect.
    \item \textbf{K and V structures are largely independent.} The Pearson correlation between K-ratio and V-ratio across all layer-step pairs is $r = -0.17$, confirming that the two spaces carry largely orthogonal structural information. This supports the theoretical prediction that the two channels provide complementary control.
\end{itemize}

\subsection{Qualitative Comparison}
\label{sec:qualitative}

\begin{figure*}[t]
\centering
\includegraphics[width=\linewidth]{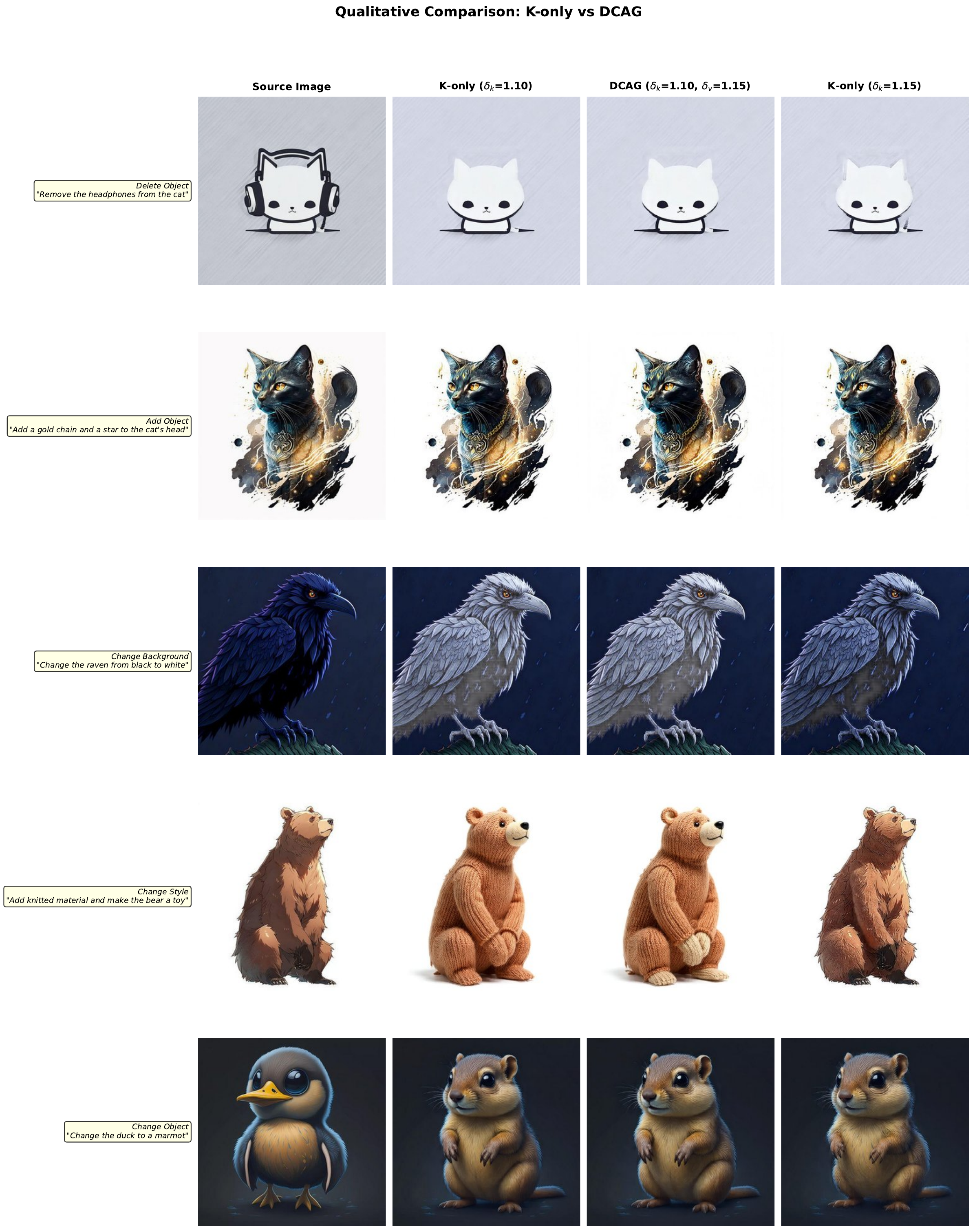}
\caption{Qualitative comparison across editing categories. DCAG ($\delta_k\!=\!1.10, \delta_v\!=\!1.15$) better preserves non-edited regions compared to K-only ($\delta_k\!=\!1.10$), while maintaining editing quality. K-only ($\delta_k\!=\!1.15$) achieves higher fidelity but at the cost of reduced editing strength.}
\label{fig:qualitative}
\end{figure*}

\cref{fig:qualitative} presents qualitative comparisons across five editing categories. For each example, we show the source image, K-only guidance ($\delta_k\!=\!1.10$), DCAG ($\delta_k\!=\!1.10, \delta_v\!=\!1.15$), and stronger K-only guidance ($\delta_k\!=\!1.15$) for reference.

%% file: discussion.tex
\section{Discussion and Conclusion}
\label{sec:discussion}

\paragraph{Why Dual-Channel Outperforms Single-Channel.}
The Key and Value channels operate on different factors of the attention output $o = \sum_i \alpha_i V^i$. Key-channel guidance modifies the weights $\{\alpha_i\}$, redistributing attention across tokens. Value-channel guidance modifies the features $\{V^i\}$, adjusting what content each token contributes. Because these are multiplicative factors, their effects compose rather than interfere. DCAG exploits this composability to achieve fidelity improvements that are inaccessible to either channel alone.

\paragraph{Practical Guidelines.}
Based on our experiments, we recommend the following guidelines for practitioners:
\begin{itemize}
    \item Use $\delta_k = 1.10, \delta_v = 1.15$ as the default configuration (best overall fidelity).
    \item At moderate Key guidance ($\delta_k \approx 1.10$), the Value channel improves nearly all edit types; $\delta_v \in [1.05, 1.15]$ is the effective range.
    \item At stronger Key guidance ($\delta_k \geq 1.15$), use the Value channel conservatively ($\delta_v \leq 1.05$) or omit it, as some categories may regress.
    \item For localized edits (delete/add objects, change background), $\delta_v$ provides the most benefit.
    \item For global edits (change action/position), $\delta_v$ has limited effect; focus on $\delta_k$.
    \item The Value channel saturates at $\delta_v \approx 1.15$; beyond this, features begin to distort rather than sharpen.
\end{itemize}

\paragraph{Limitations.}
The Value channel's effect is inherently milder than the Key channel due to its linear nature. While this makes it predictable and safe (no risk of artifacts), the overall improvement is modest (1.8\% LPIPS at best). Furthermore, the Value channel exhibits diminishing returns at higher $\delta_k$: at $\delta_k = 1.15$, the improvement drops to 0.9\% and some categories regress. This interaction between channels suggests that the two are not fully independent in practice—strong Key-channel guidance already captures most achievable fidelity gains, leaving less room for Value-channel contribution. Our current evaluation focuses on fidelity metrics; a comprehensive study of editing quality trade-offs (e.g., via CLIP-Score) across the 2D parameter space remains future work.

\paragraph{Future Directions.}
Several extensions are worth exploring: (1) \emph{Spatially-adaptive DCAG}, where $(\delta_k, \delta_v)$ vary per-token based on edit relevance, similar to mask-guided approaches~\cite{li2024commerce,zhang2023adding}; (2) \emph{Query-space guidance}, extending the bias-delta framework to the third projection; (3) \emph{Video editing}, where temporal consistency adds another dimension to the control problem; (4) integration with identity-preserving generation~\cite{li2025editid,li2025editidv2,li2026flexid} for controlled personalized editing.

\paragraph{Conclusion.}
We presented Dual-Channel Attention Guidance (DCAG), a training-free framework for fine-grained image editing control in Diffusion Transformers. By discovering and exploiting the bias-delta structure in both Key and Value projections, DCAG provides a 2D parameter space that enables more precise editing-fidelity trade-offs than single-channel methods. Our theoretical analysis reveals the complementary nature of the two channels—Key for coarse nonlinear control, Value for fine linear control—and extensive experiments on PIE-Bench validate consistent improvements across diverse editing categories.

\section*{Declarations}

\subsection*{Funding and/or Conflicts of Interests/Competing Interests}
The authors declare that they have no known competing financial interests or personal relationships that could have appeared to influence the work reported in this paper. The authors did not receive support from any organization for the submitted work.

\subsection*{Ethical and Informed Consent for Data Used}
This article does not contain any studies with human participants or animals performed by any of the authors. The research utilizes publicly available datasets (PIE-Bench~\cite{ju2023pnp}) and pre-trained models. Therefore, ethical approval and informed consent were not required.

\subsection*{Author Contribution}
Guandong Li proposed the original idea, designed the dual-channel attention guidance framework, conducted all experiments, and wrote the manuscript. Mengxia Ye contributed to discussions on methodology and provided feedback on the manuscript.

\subsection*{Data Availability}
The datasets used in this study are available in the PIE-Bench repository (\url{https://github.com/timothybrooks/instruct-pix2pix}). The pre-trained base model (Qwen-Image-Edit) used in this study is publicly available from HuggingFace (\url{https://huggingface.co/Qwen/Qwen-Image-Edit}). The evaluation metrics rely on publicly available implementations of LPIPS~\cite{zhang2018unreasonable}, SSIM~\cite{wang2004image}, and CLIP~\cite{radford2021learning}.

%% file: references.bib
@article{zhang2025group,
  title={Group Relative Attention Guidance for Image Editing},
  author={Zhang, Xuanpu and Niu, Xuesong and Chen, Ruidong and Song, Dan and Zeng, Jianhao and Du, Penghui and Cao, Haoxiang and Wu, Kai and Liu, An-an},
  journal={arXiv preprint arXiv:2510.24657},
  year={2025}
}

@article{li2024layout,
  title={Layout control and semantic guidance with attention loss backward for t2i diffusion model},
  author={Li, Guandong},
  journal={arXiv preprint arXiv:2411.06692},
  year={2024}
}

@article{li2024commerce,
  title={E-Commerce Inpainting with Mask Guidance in Controlnet for Reducing Overcompletion},
  author={Li, Guandong},
  journal={arXiv preprint arXiv:2409.09681},
  year={2024}
}

@inproceedings{li2025editid,
  title={Editid: Training-free editable id customization for text-to-image generation},
  author={Li, Guandong and Chu, Zhaobin},
  booktitle={Findings of the Association for Computational Linguistics: EMNLP 2025},
  pages={6301--6319},
  year={2025}
}

@article{li2025editidv2,
  title={EditIDv2: Editable ID Customization with Data-Lubricated ID Feature Integration for Text-to-Image Generation},
  author={Li, Guandong and Chu, Zhaobin},
  journal={arXiv preprint arXiv:2509.05659},
  year={2025}
}

@article{li2026flexid,
  title={FlexID: Training-Free Flexible Identity Injection via Intent-Aware Modulation for Text-to-Image Generation},
  author={Li, Guandong and Ding, Yijun},
  journal={arXiv preprint arXiv:2602.07554},
  year={2026}
}

@article{ho2020denoising,
  title={Denoising diffusion probabilistic models},
  author={Ho, Jonathan and Jain, Ajay and Abbeel, Pieter},
  journal={Advances in neural information processing systems},
  volume={33},
  pages={6840--6851},
  year={2020}
}

@inproceedings{rombach2022high,
  title={High-resolution image synthesis with latent diffusion models},
  author={Rombach, Robin and Blattmann, Andreas and Lorenz, Dominik and Esser, Patrick and Ommer, Bj{\"o}rn},
  booktitle={Proceedings of the IEEE/CVF conference on computer vision and pattern recognition},
  pages={10684--10695},
  year={2022}
}

@article{song2020denoising,
  title={Denoising diffusion implicit models},
  author={Song, Jiaming and Meng, Chenlin and Ermon, Stefano},
  journal={arXiv preprint arXiv:2010.02502},
  year={2020}
}

@article{lipman2022flow,
  title={Flow matching for generative modeling},
  author={Lipman, Yaron and Chen, Ricky TQ and Ben-Hamu, Heli and Nickel, Maximilian and Le, Matt},
  journal={arXiv preprint arXiv:2210.02747},
  year={2022}
}

@article{liu2022flow,
  title={Flow straight and fast: Learning to generate and transfer data with rectified flow},
  author={Liu, Xingchao and Gong, Chengyue and Liu, Qiang},
  journal={arXiv preprint arXiv:2209.03003},
  year={2022}
}

@article{ho2022classifier,
  title={Classifier-free diffusion guidance},
  author={Ho, Jonathan and Salimans, Tim},
  journal={arXiv preprint arXiv:2207.12598},
  year={2022}
}

@inproceedings{peebles2023scalable,
  title={Scalable diffusion models with transformers},
  author={Peebles, William and Xie, Saining},
  booktitle={Proceedings of the IEEE/CVF international conference on computer vision},
  pages={4195--4205},
  year={2023}
}

@article{su2024roformer,
  title={Roformer: Enhanced transformer with rotary position embedding},
  author={Su, Jianlin and Ahmed, Murtadha and Lu, Yu and Pan, Shengfeng and Bo, Wen and Liu, Yunfeng},
  journal={Neurocomputing},
  volume={568},
  pages={127063},
  year={2024},
  publisher={Elsevier}
}

@inproceedings{brooks2023instructpix2pix,
  title={Instructpix2pix: Learning to follow image editing instructions},
  author={Brooks, Tim and Holynski, Aleksander and Efros, Alexei A},
  booktitle={Proceedings of the IEEE/CVF conference on computer vision and pattern recognition},
  pages={18392--18402},
  year={2023}
}

@inproceedings{mokady2023null,
  title={Null-text inversion for editing real images using guided diffusion models},
  author={Mokady, Ron and Hertz, Amir and Aberman, Kfir and Pritch, Yael and Cohen-Or, Daniel},
  booktitle={Proceedings of the IEEE/CVF conference on computer vision and pattern recognition},
  pages={6038--6047},
  year={2023}
}

@inproceedings{tumanyan2023plug,
  title={Plug-and-play diffusion features for text-driven image-to-image translation},
  author={Tumanyan, Narek and Geyer, Michal and Bagon, Shai and Dekel, Tali},
  booktitle={Proceedings of the IEEE/CVF conference on computer vision and pattern recognition},
  pages={1921--1930},
  year={2023}
}

@inproceedings{kawar2023imagic,
  title={Imagic: Text-based real image editing with diffusion models},
  author={Kawar, Bahjat and Zada, Shiran and Lang, Oran and Tov, Omer and Chang, Huiwen and Dekel, Tali and Mosseri, Inbar and Irani, Michal},
  booktitle={Proceedings of the IEEE/CVF conference on computer vision and pattern recognition},
  pages={6007--6017},
  year={2023}
}

@article{meng2021sdedit,
  title={Sdedit: Guided image synthesis and editing with stochastic differential equations},
  author={Meng, Chenlin and He, Yutong and Song, Yang and Song, Jiaming and Wu, Jiajun and Zhu, Jun-Yan and Ermon, Stefano},
  journal={arXiv preprint arXiv:2108.01073},
  year={2021}
}

@inproceedings{kulikov2025flowedit,
  title={Flowedit: Inversion-free text-based editing using pre-trained flow models},
  author={Kulikov, Vladimir and Kleiner, Matan and Huberman-Spiegelglas, Inbar and Michaeli, Tomer},
  booktitle={Proceedings of the IEEE/CVF International Conference on Computer Vision},
  pages={19721--19730},
  year={2025}
}

@article{hertz2022prompt,
  title={Prompt-to-prompt image editing with cross attention control},
  author={Hertz, Amir and Mokady, Ron and Tenenbaum, Jay and Aberman, Kfir and Pritch, Yael and Cohen-Or, Daniel},
  journal={arXiv preprint arXiv:2208.01626},
  year={2022}
}

@article{chefer2023attend,
  title={Attend-and-excite: Attention-based semantic guidance for text-to-image diffusion models},
  author={Chefer, Hila and Alaluf, Yuval and Vinker, Yael and Wolf, Lior and Cohen-Or, Daniel},
  journal={ACM transactions on Graphics (TOG)},
  volume={42},
  number={4},
  pages={1--10},
  year={2023},
  publisher={ACM New York, NY, USA}
}

@article{epstein2023diffusion,
  title={Diffusion self-guidance for controllable image generation},
  author={Epstein, Dave and Jabri, Allan and Poole, Ben and Efros, Alexei and Holynski, Aleksander},
  journal={Advances in Neural Information Processing Systems},
  volume={36},
  pages={16222--16239},
  year={2023}
}

@inproceedings{cao2023masactrl,
  title={Masactrl: Tuning-free mutual self-attention control for consistent image synthesis and editing},
  author={Cao, Mingdeng and Wang, Xintao and Qi, Zhongang and Shan, Ying and Qie, Xiaohu and Zheng, Yinqiang},
  booktitle={Proceedings of the IEEE/CVF international conference on computer vision},
  pages={22560--22570},
  year={2023}
}

@inproceedings{ahn2024self,
  title={Self-rectifying diffusion sampling with perturbed-attention guidance},
  author={Ahn, Donghoon and Cho, Hyoungwon and Min, Jaewon and Jang, Wooseok and Kim, Jungwoo and Kim, SeonHwa and Park, Hyun Hee and Jin, Kyong Hwan and Kim, Seungryong},
  booktitle={European Conference on Computer Vision},
  pages={1--17},
  year={2024},
  organization={Springer}
}

@article{qwen_image_edit,
  title={Qwen-Image-Edit},
  author={{Qwen Team}},
  howpublished={\url{https://huggingface.co/Qwen/Qwen-Image-Edit}},
  year={2025}
}

@article{liu2025step1x,
  title={Step1x-edit: A practical framework for general image editing},
  author={Liu, Shiyu and Han, Yucheng and Xing, Peng and Yin, Fukun and Wang, Rui and Cheng, Wei and Liao, Jiaqi and Wang, Yingming and Fu, Honghao and Han, Chunrui and others},
  journal={arXiv preprint arXiv:2504.17761},
  year={2025}
}

@inproceedings{ju2023pnp,
  title={Pnp inversion: Boosting diffusion-based editing with 3 lines of code},
  author={Ju, Xuan and Zeng, Ailing and Bian, Yuxuan and Liu, Shaoteng and Xu, Qiang},
  booktitle={The Twelfth International Conference on Learning Representations},
  year={2023}
}

@inproceedings{zhang2018unreasonable,
  title={The unreasonable effectiveness of deep features as a perceptual metric},
  author={Zhang, Richard and Isola, Phillip and Efros, Alexei A and Shechtman, Eli and Wang, Oliver},
  booktitle={Proceedings of the IEEE conference on computer vision and pattern recognition},
  pages={586--595},
  year={2018}
}

@article{wang2004image,
  title={Image quality assessment: from error visibility to structural similarity},
  author={Wang, Zhou and Bovik, Alan C and Sheikh, Hamid R and Simoncelli, Eero P},
  journal={IEEE transactions on image processing},
  volume={13},
  number={4},
  pages={600--612},
  year={2004},
  publisher={IEEE}
}

@inproceedings{radford2021learning,
  title={Learning transferable visual models from natural language supervision},
  author={Radford, Alec and Kim, Jong Wook and Hallacy, Chris and Ramesh, Aditya and Goh, Gabriel and Agarwal, Sandhini and Sastry, Girish and Askell, Amanda and Mishkin, Pamela and Clark, Jack and others},
  booktitle={International conference on machine learning},
  pages={8748--8763},
  year={2021},
  organization={PmLR}
}

@inproceedings{zhang2023adding,
  title={Adding conditional control to text-to-image diffusion models},
  author={Zhang, Lvmin and Rao, Anyi and Agrawala, Maneesh},
  booktitle={Proceedings of the IEEE/CVF international conference on computer vision},
  pages={3836--3847},
  year={2023}
}

@article{podell2023sdxl,
  title={Sdxl: Improving latent diffusion models for high-resolution image synthesis},
  author={Podell, Dustin and English, Zion and Lacey, Kyle and Blattmann, Andreas and Dockhorn, Tim and M{\"u}ller, Jonas and Penna, Joe and Rombach, Robin},
  journal={arXiv preprint arXiv:2307.01952},
  year={2023}
}

@inproceedings{esser2024scaling,
  title={Scaling rectified flow transformers for high-resolution image synthesis},
  author={Esser, Patrick and Kulal, Sumith and Blattmann, Andreas and Entezari, Rahim and M{\"u}ller, Jonas and Saini, Harry and Levi, Yam and Lorenz, Dominik and Sauer, Axel and Boesel, Frederic and others},
  booktitle={Forty-first international conference on machine learning},
  year={2024}
}

@article{dhariwal2021diffusion,
  title={Diffusion models beat gans on image synthesis},
  author={Dhariwal, Prafulla and Nichol, Alexander},
  journal={Advances in neural information processing systems},
  volume={34},
  pages={8780--8794},
  year={2021}
}
